\documentclass[letterpaper]{article} 
\usepackage{aaai23}  
\usepackage{times}  
\usepackage{helvet}  
\usepackage{courier}  
\usepackage[hyphens]{url}  
\usepackage{graphicx} 
\usepackage{natbib}  
\usepackage{caption} 
\usepackage{algorithm}
\usepackage{algorithmic}

\usepackage{newfloat}
\usepackage{listings}
\DeclareCaptionStyle{ruled}{labelfont=normalfont,labelsep=colon,strut=off} 
\floatstyle{ruled}
\newfloat{listing}{tb}{lst}{}
\floatname{listing}{Listing}
\title{SciCap+: A Knowledge Augmented Dataset to \\Study the Challenges of Scientific Figure Captioning}
\author{
    Zhishen Yang\textsuperscript{\rm 1}, Raj Dabre\textsuperscript{\rm 2}, Hideki Tanaka\textsuperscript{\rm 2}, Naoaki Okazaki\textsuperscript{\rm 1}\\
}
\affiliations{
    \textsuperscript{\rm 1} Tokyo Institute of Technology\\2-12-1 Ookayama, Meguro-ku, Tokyo, 152-8550, Japan \\
    \textsuperscript{\rm 2} National Institute of Information and Communications Technology\\3-5 Hikaridai, Seika-cho, Soraku-gun, Kyoto, 619-0289, Japan\\
    zhishen.yang@nlp.c.titech.ac.jp, raj.dabre@nict.go.jp, hideki.tanaka@nict.go.jp,  okazaki@c.titech.ac.jp
}

\usepackage{bibentry}

\begin{document}

\maketitle

\begin{abstract}
In scholarly documents, figures provide a straightforward way of communicating scientific findings to readers. Automating figure caption generation helps move model understandings of scientific documents beyond text and will help authors write informative captions that facilitate communicating scientific findings. Unlike previous studies, we reframe scientific figure captioning as a knowledge-augmented image captioning task that models need to utilize knowledge embedded across modalities for caption generation. To this end, we extended the large-scale SciCap dataset~\cite{hsu-etal-2021-scicap-generating} to SciCap+ which includes mention-paragraphs (paragraphs mentioning figures) and OCR tokens. Then, we conduct experiments with the M4C-Captioner (a multimodal transformer-based model with a pointer network) as a baseline for our study. Our results indicate that mention-paragraphs serves as additional context knowledge, which significantly boosts the automatic standard image caption evaluation scores compared to the figure-only baselines. Human evaluations further reveal the challenges of generating figure captions that are informative to readers. The code and SciCap+ dataset will be publicly available:\footnote{\url{https://github.com/ZhishenYang/scientific_figure_captioning_dataset}}

\end{abstract}

\section{Introduction}

Scholarly documents are the primary source for sharing scientific knowledge. These documents are available in various formats, such as journal articles, book chapters, and conference proceedings. A significant portion of these documents is text and together with figures and tables, they help communicate knowledge to readers. Using figures provides visual representations of complex information that facilitate the sharing of scientific findings with readers efficiently and straightforwardly. The standard practice for scientific writing is to write a caption for each figure, accompanied by paragraphs with detailed explanations. Figures and captions should be standalone, and readers should be able to understand the figures without referring to the main text. Helping authors write appropriate and informative captions for figures will improve the quality of scientific documents, thereby enhancing the speed and quality of scientific communication. In this study, we focus on automating the generation of captions for figures in scientific papers.

Scientific figure captioning is a variant of the image captioning task. However, with the same goal of generating a caption, it has two unique challenges: 1. Figures are not natural images: In contrast to natural images, visual objects are texts and data points in scientific figures. 2. The captions of the figures should explain: Instead of simply identifying objects and texts in the figures, the caption should contain an analysis that the authors intend to present and highlight findings. 

A previous study~\cite{hsu-etal-2021-scicap-generating}, SciCap, defines the scientific figure captioning task as a figure-to-caption task: A model generates captions only referring to figures. Their work reported relatively lower scores as measured by automatic evaluation metrics, indicating that there is considerable room for improvement. Intuitively, writing appropriate figure captions without sufficient background knowledge is difficult, since even humans will struggle to interpret a figure and write a caption unless some background knowledge is available. On the basis of this observation, we think that generating appropriate captions is infeasible without adding context knowledge to the caption generation model. This context comes in two forms: background knowledge from the running text and the OCR tokens in the figure, both of which should help reduce the burden on the captioning model. To this end, we augment the existing large-scale scientific figure captioning dataset: SciCap with mention-paragraphs and OCR tokens and call the resultant dataset as SciCap+. We then pose scientific figure captioning as a multimodal summarization task and use the M4C captioner model~\cite{sidorov2020textcaps}(a model that utilizes multimodal knowledge to generate captions) as a baseline to study the scientific figure captioning task. The experimental result of automatic evaluation demonstrates that using knowledge embedded in different modalities, especially in the form of mention-paragraphs and OCR tokens, significantly boosts performance. 

In addition to experiments using automatic evaluation metrics, we also performed human generation and evaluation tasks in order to establish the inherent difficulty of scientific figure captioning. The results of the human evaluation reveal three findings: 1. Multimodal knowledge helps models outperform humans in caption generation tasks. 2. Model-generated captions are almost as informative as ground-truth captions: Human evaluators do not prefer either type of caption. 3. Even referring to mention-paragraphs, it is still challenging for humans to write captions that are close to ground truth. To the best of our knowledge, we are the first to pose scientific figure captioning as a multimodel summarization task and show that mention-paragraphs and OCR tokens as context substantially enhance the quality of generated captions. 

\section{Preliminary Study}

\begin{figure}[tb]
    \centering
        \includegraphics[width=0.5\textwidth]{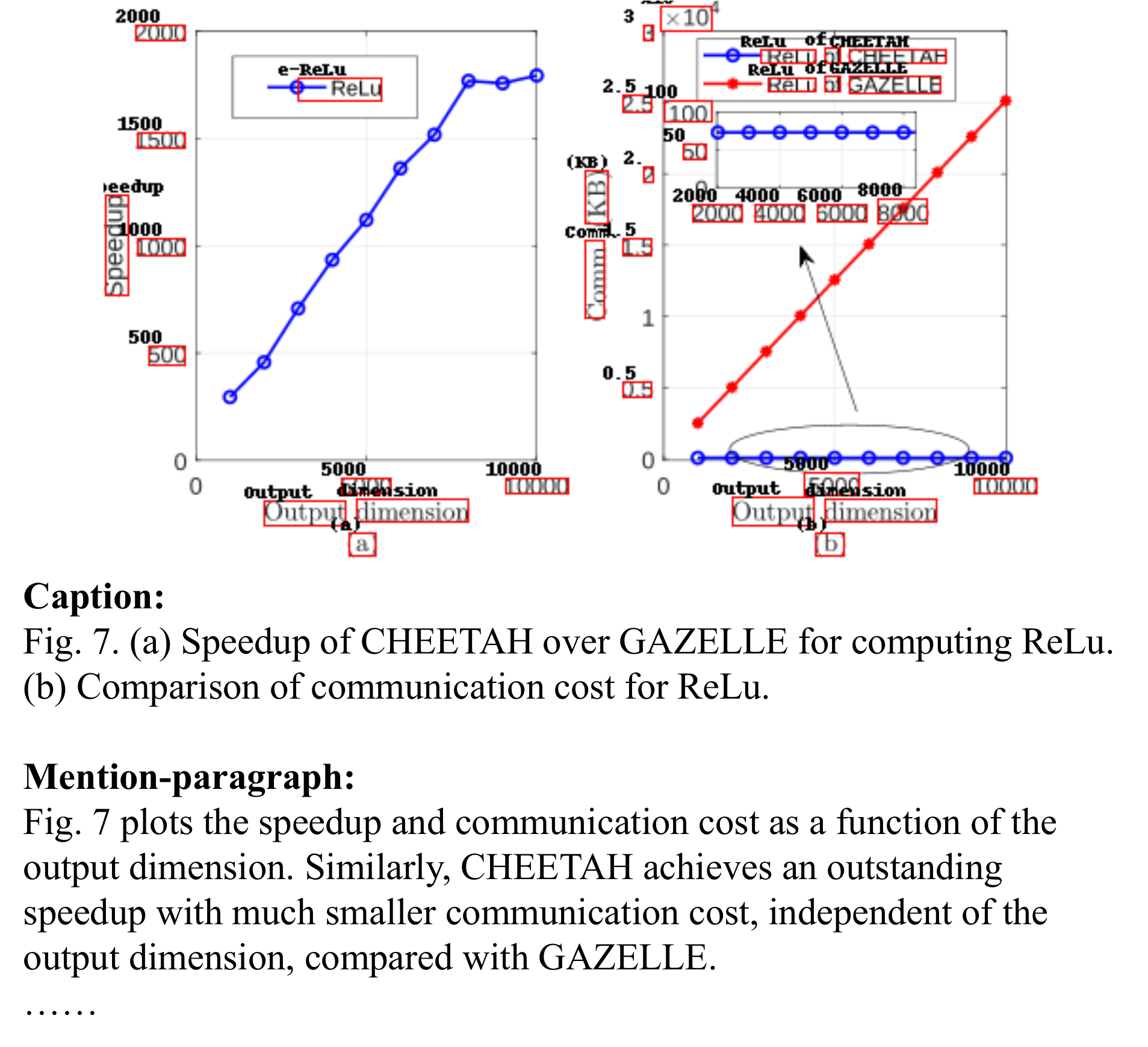} 
    \caption{An example figure~\cite{zhang2019cheetah} with its captions and mention-paragraph and the text tokens recognized via OCR. Without referring to the mention-paragraph and the OCR tokens to tie the figure and the mention, we cannot have a proper interpretation of the data presented in the figure, which is communication cost comparison and speed up of CHEETAH over GAZELLE.}  
    \label{pre_study}
\end{figure}

In the traditional image captioning task, captioning an image aims at describing the appearances or natures of recognized objects and illustrating the relationships between recognized objects. Unlike the usual image captioning tasks, figures do not contain visual scenes. Instead, the captions provide interpretations of data presented in figures to highlight scientific findings that authors want to present to readers. With this unique characteristic, without referring to mention-paragraphs, which usually refer to the figure, it is extremely challenging for a human to have proper interpretations of figures. This is because they may lack background knowledge of the domain or context of the figure. As figure~\ref{pre_study} shows, by only looking at the figure, we do not know what "comm.(KB)" stands for; therefore lacking the knowledge to write informative captions is challenging. However, the mention-paragraph contains "communication cost" and this is also present in the caption, indicating that such background knowledge should help in writing accurate captions. 

\section{Problem Formulation}
The previous study~\cite{hsu-etal-2021-scicap-generating} defined this task as an image captioning task as: Given a figure $I$, the model generates a caption $C=[c_0,c_1,...,c_N]$. However, we reframe the scientific figure captioning task as a knowledge-augmented image captioning task requiring knowledge extracted from text and vision modalities. For a figure, we define a paragraph that mentions it (mention-paragraph) and text within the figure, extracted via OCR, as text modalities. The figure itself and visual appearances of OCR texts are visual modalities. Given a scientific figure $I$ and knowledge extracted from text and vision modality: $K_{text}$ and $K_{vision}$, we define figure caption generation task as $P(C|I,K_{text}, K_{vision})$ 


\section{SciCap+ Dataset}

\begin{figure*}[tb]
    \centering
        \includegraphics[width=\textwidth]{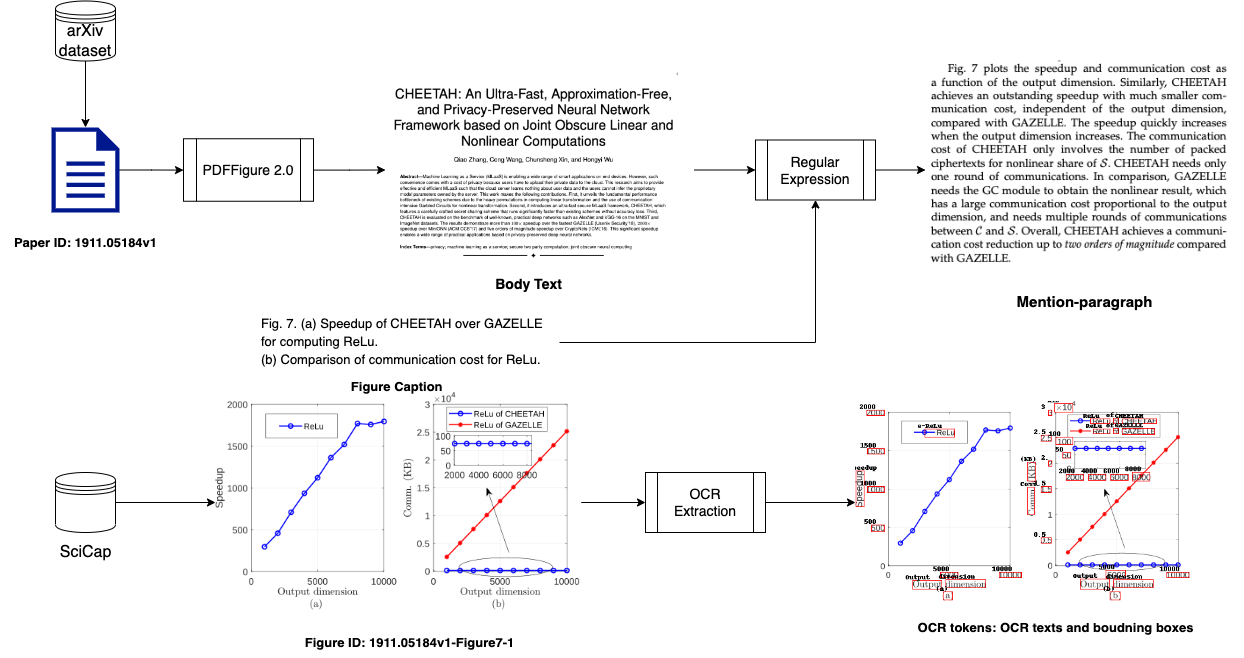} 
    \caption{The overall workflow of the data augmentation for creating SciCap+ dataset. For each figure in SciCap+, we extracted its mention-paragraphs and  OCR tokens (OCR texts and bounding boxes).     }  
    \label{dataset_creation}
\end{figure*}

SciCap is a large-scale figure-caption dataset comprising graph plots extracted from 10 years of collections of arXiv computer science papers. We used around 414k figures from SciCap and augment each figure with its mention-paragraphs and OCR tokens with metadata. This section details the data set creation and data augmentation processes. Figure~\ref{dataset_creation} shows the overall workflow behind the creation of SciCap+.

\subsection{Mention-paragraph Extraction}

We first obtained papers in PDF format from Kaggle arXiv dastaset~\footnote{\url{https://www.kaggle.com/datasets/Cornell-University/arxiv}}. The reason for using PDFs is that not all papers have source files and some are complicated to parse. 
After obtaining PDFs, we used PDFFigures 2.0~\cite{clark2016pdffigures}~\footnote{\url{https://github.com/allenai/pdffigures2}} to extract the body text of each paper. PDFFigure 2.0 is a tool that extracts figures, captions, tables, and text from scholarly PDFs in computer science. In scholarly documents, authors label figures with numbers (e.g. Figure 1. Fig. 1). For a figure, we used its figure number in a regular expression to locate a paragraph that mentions it. 

\subsection{OCR Extraction}

The SciCap dataset also provides texts extracted from figures as metadata, but does not provide location information for each text. To include location information for each text in a figure, we used Google Vision OCR API to extract text tokens from each figure with its coordinates of bounding boxes.

\subsection{Data Statistics}

The splitting of the SciCap dataset is at the figure level. Therefore, figures from the same paper may appear in different splits. This will lead to unfair evaluation, since the information of one figure in one split may coincidentally overlap with the information of another figure.  We thus re-split figures at the document level to eliminate this overlapping problem. 
\citet{hsu-etal-2021-scicap-generating} show that text normalization and figure filtering do not improve model performance. Hence, we keep original captions and all figures (with/without sub-figures) in the SciCap+ dataset. For a figure, we kept only the first paragraph that mentions it in the body text. Table\ref{data_stats} shows statistics of the SciCap+ dataset. In all three splits, around 90\% of the captions are less than 66 words. All figures are graph plots.

\begin{table}[t]
    \centering
    \begin{tabular}{lrr}
        \hline
        \textbf{Split} & \textbf{Figures} &  \textbf{Words} \\
        \hline
        Training & {394,005} & {12,336,511} \\
        Test & {10,336} & {323,382} \\
        Validation & {10,468} & {329,072} \\
        \hline
\end{tabular}

\caption{Statistics of the SciCap+ dataset.}
\label{data_stats}
\end{table}

\subsection{Dataset Quality Evaluation} 
Before conducting experiments, we conducted human evaluation of SciCap+ where we checked the mention-paragraphs and OCR tokens extraction quality. The aim was to establish whether the mention-paragraphs and OCR tokens were extracted correctly and relevant to the figure and its caption. To this end, we randomly selected 200 figures from the training set and for each figure, we asked two human evaluators to give scores of 1-5 (1 represents no relevance and 5 is highly relevant) for relevance between a caption of a figure and its mention-paragraphs and OCR tokens. 

Compared to natural image captioning, human evaluation tasks for the figure captioning domain requires expert knowledge. We recruited two colleagues to carry out this evaluation task. Both of them have Ph.D. degrees in computer science and work as researchers. Their experience implies that they have adequate experience writing figure captions. 

\begin{figure}[tb]
    \centering
        \includegraphics[width=0.45\textwidth]{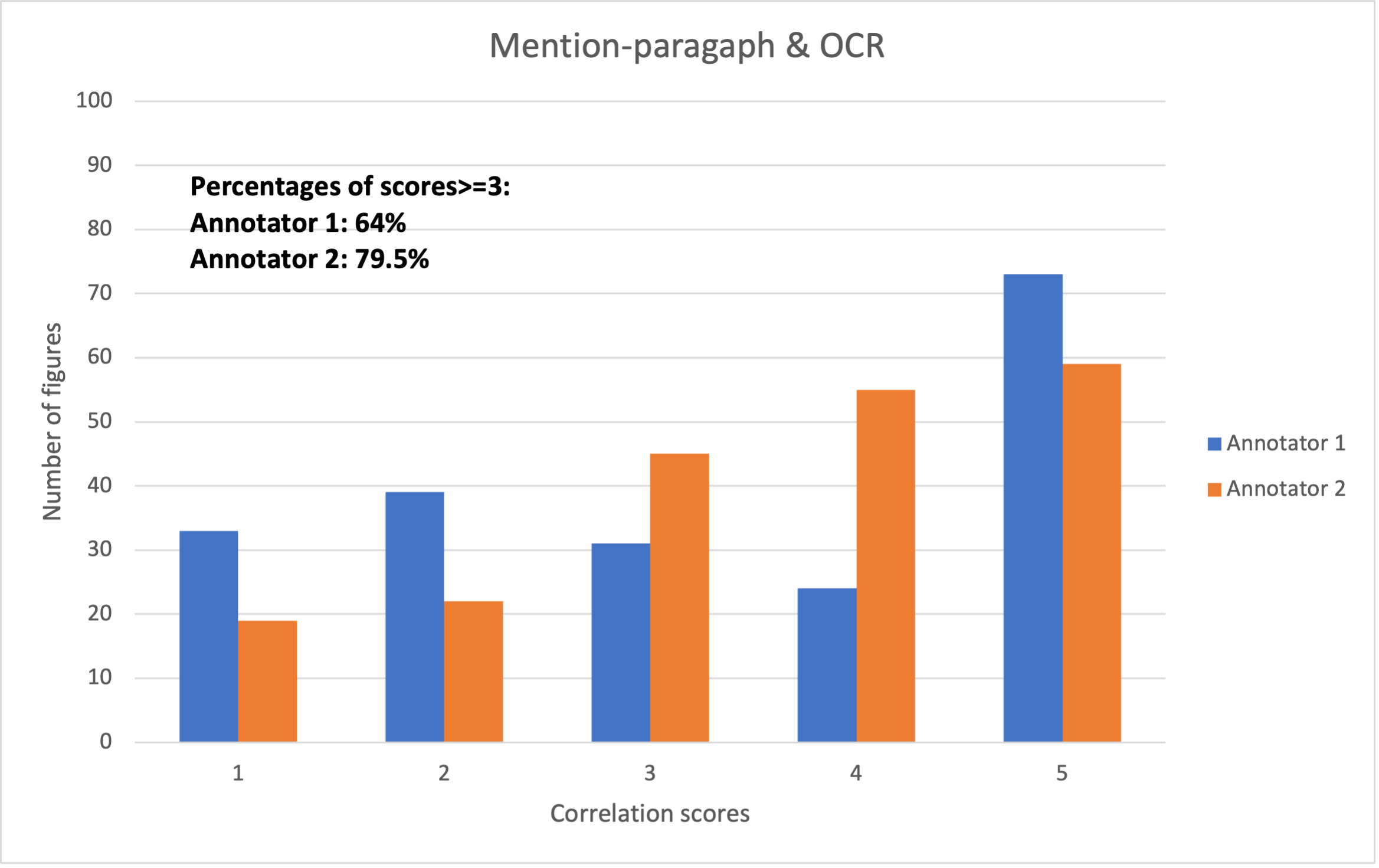} 
    \caption{Score distribution on correlations between mention-paragraph, OCR tokens and figure captions. Both evaluators judged most of the figures with at least moderate correlations with captions. }  
    \label{dataset_quality_check}
\end{figure}

Figure~\ref{dataset_quality_check} shows the distributions of the relevance scores. We can observe that two evaluators gave most of the figures (evaluator 1: 64\% and evaluator 2: 79.5\%) with relevance scores greater than 3 and a cohen kappa score of 0.28. This evaluation result indicates that the mention-paragraphs and OCR tokens have a satisfactory extraction quality and that the annotators considered most of them as relevant to the figure and its caption. However, the two annotators seem to have a relatively lower agreement (0.28) regarding which figures and captions are relevant to their mention-paragraphs and OCR tokens. We attribute this to the fact that evaluations of figure captions are highly subjective.

\section{Experiments}

We conduct experiments using SciCap+ to empirically prove that scientific figure captioning is inherently a knowledge-augmented task and benefits from knowledge coming from both text and vision modalities. 

\subsection{Figure Captioning Model}

We used M4C-Captioner~\cite{sidorov2020textcaps} as the baseline model to study the scientific figure captioning task. The M4C-Captioner is based on Multimodal Multi-Copy Mesh (M4C)~\cite{hu2020iterative} that jointly learns representations across input modalities. To solve the out-of-vocabulary problem during caption generation, it is equipped with a pointer network that picks up text from OCR tokens or a predefined fixed dictionary. In this work, 3 input features are used, figure, mention-paragraphs and OCR tokens fed to encoders, the output representations of which are fed to the M4C-Captioner.

\subsection{Implementation and Training}

Our implementation of M4C-Captioner is based on the MMF framework~\cite{singh2020mmf} and Pytorch. The implementation allows users to specify diverse pre-trained encoders for each modality, which can be fine-tuned or frozen during training. The M4C-captioner itself has $D=768$  hidden dimension size, $K=4$ transformer layers and 12 attention heads. We used sentencepiece~\cite{kudo-richardson-2018-sentencepiece} to obtain a dictionary of 32000 subwords built from both mention-paragraphs and OCR tokens. This is used as the M4C-captioner's vocabulary. We followed the BERT-BASE hyperparameter setting and trained from scratch.

Regarding the encoders that feed features to M4C-captioner, we used pre-trained Resnet-152 as the figure's vision encoder. For each figure, we applied a 2D adaptive average pooling over outputs from layer 5 to obtain a global visual feature vector with a dimension of 2048. Layers 2, 3 and 4 layers were fine-tuned during training. For mention-paragraph features, SciBERT~\cite{beltagy-etal-2019-scibert} was used to encode\footnote{We only used the first 3 layers of SciBERT for lightweightness.} it into 758-dimensional feature vectors. The number of vectors equals the number of sub-word tokens in the mention-paragraph, which we limit to 192. The mention-paragraph encoder is also fine-tuned during training.  Finally, for OCR tokens, we use both text and visual features. We selected FastText~\cite{almazan2014word} as the word encoder and Pyramidal Histogram of Characters (PHOC)~\cite{bojanowski2017enriching} as the character encoder. Regarding the visual feature encoder of OCR tokens, we first extracted Faster R-CNN fc6 features and then applied fc7 weights to it to obtain 2048-dimensional appearance features for bounding boxes of OCR tokens. The fc7 weights were fine-tuned during training. We kept a maximum of 95 OCR tokens per figure.  

We trained a model on a GPU server with 8 Nvidia Tesla V100 GPUs. Training a model with a complete set of features took 13 hours. During training, we used a batch size of 128. We selected CIDEr as the evaluation metric. The evaluation interval is every 2000 iterations, we stop training if CIDEr score does not improve for 4 evaluation intervals. The optimizer is Adam with a learning rate of $0.001$ and $\epsilon= 1.0\mathrm{E}{-08} $. We also used a multistep learning rate schedule with warmup iterations of 1000 and a warmup factor of 0.2. We kept the maximum number of decoding steps at the decoding time as 67. 
For evaluation, we used five standard metrics for evaluating image captions: BLEU-4~\cite{papineni-etal-2002-bleu}, METEOR~\cite{banerjee2005meteor}, ROUGE-L~\cite{lin2004rouge}, CIDEr~\cite{vedantam2015cider} and SPICE~\cite{anderson2016spice}. Since figure captions contain scientific terms which can be seen as uncommon words, among all five metrics, we are particularly interested in CIDEr since it emphasizes them.






\begin{table*}
    \centering
    \resizebox{\linewidth}{!}{
    \begin{tabular}{lrrrrrr}
    \hline
    \textbf{Model} & \textbf{BLEU-4} & \textbf{METEOR} & \textbf{ROUGE-L} & \textbf{SPICE} & \textbf{CIDEr}  \\
    \hline
    1. M4C-Captioner (Figure Only ) & 1.5 & 5.6 & 15.4 & 4.3 & 4.6 \\
    2. M4C-Captioner (Mention Only)& 5.3 & 11.0 & 27.4 & 14.3 & 49.0 \\
    3. M4C-Captioner (Figure and OCR features) & 2.6 & 7.6 & 20.5 & 10.1 & 22.2  \\
    4. M4C-Captioner (Mention, Figure and OCR features)  & 6.3 & \textbf{12.0} & 29.2 &  15.8 & 55.8 \\
    \hline
     \textbf{Ablation Study on Figures} \\
    \hline
    5. M4C-Captioner (Mention and OCR features) & 6.3 & \textbf{12.0} & \textbf{29.3} & \textbf{16.1} & \textbf{56.4} \\

    \hline
     \textbf{Ablation Study on OCR features} \\
    \hline
    6. M4C-Captioner (Mention, Figure and w/o OCR features )  & \textbf{6.4} & 11.5 & 27.9 & 14.6 & 50.5 \\
    7. M4C-Captioner (Mention, Figure and OCR spatial features) & 5.8  & 11.1 & 27.3 & 14.1 & 48.0 \\
    
    8. M4C-Captioner (Mention, Figure and OCR (w/o spatial features) features )  & \textbf{6.4} & \textbf{12.0} & 29.1 & 15.7 & 54.6 \\
    9. M4C-Captioner (Mention, Figure and OCR (w/o visual features) features )  & 6.2 & 11.9 & 28.9 & 15.6 & 54.1 \\
    \hline
    \end{tabular}   
    }
    
    \caption{Automatic evaluation scores of M4C-captioning on SciCap dataset. Aggregate knowledge from text and vision modalities significantly boosts the model performance compared to the figure-only baseline.}\label{main_result}
\end{table*}

\section{Results}

\subsection{Main Result}

The experimental results in table~\ref{main_result} demonstrate that using the mention-paragraph and OCR tokens significantly improves scores on all five metrics compared to the figure-only baseline. The experimental results align with our hypothesis and preliminary study that scientific figure captioning is a knowledge-augmented image captioning task, OCR tokens and knowledge embedded in mention-paragraphs help in composing informative captions.

We established a baseline M4C-Captioner (Figure only) with figures as the only input modality to the M4C-Captioner model in row \#1. This baseline is in the non-knowledge setting. Therefore, low scores in all metrics show that the model needs knowledge of other modalities. Using the mention only in row \#2 shows that the mention certainly contains a lot of useful information, as evidenced by the increase in performance. When OCR features are added to the figure input in row \#3, scores for all metrics have significant gains compared to the figure-only baseline, but are still weaker than when only mentions are used. This motivates the combination of mentions and OCR features and in row \#4, compared to the figure-only baseline and figure-OCR-only baseline, the performance further improves. Perhaps the most interesting result is in row \#5 where we only use the mentions and OCR features but not the figure and get the best performance, particularly for SPICE and CIDEr, albeit comparable to when the figure is included in row \#4.  All these results indicate that explicitly extracted multimodal knowledge helps to compose informative captions. 


\subsection{Ablation Studies}
We first performed an ablation study on figures by removing visual feature vectors, the CIDEr score increases slightly, indicating that the visual feature is more like noise for the model. This is likely because the Resnet-152 visual encoder we used was not trained on figures.

We enriched the representations of the OCR features by adding text, visual, and spatial features. Ablation studies aim to reveal impacts of each OCR token feature. All comparisons are with row \#4 even though row \#5 gives slightly better scores. With OCR features completely removed in row \#6, the CIDEr scores decrease by 5.3. Using only OCR spatial features in row \#7, the CIDEr score dropped by 7.8. Removing OCR spatial features in row \#8, the CIDEr scores dropped by 1.2. Upon removal of OCR visual features in row \#9, the CIDEr score is close to removing spatial features. 

The above ablation study indicates that the enriched OCR contributes to the informativeness of generated captions. Unlike OCR features, where appearance features are helpful to the model, removing visual features of figures increases CIDEr scores, further indicating that we need a specific vision encoder for figures to provide meaningful features. 

\section{Human Evaluation}\label{sec:humaneval}

Having established that knowledge helps a model perform figure captioning, we conducted some human evaluation activities to determine their subjective quality.
We conducted human caption generation and evaluation tasks. The human generation task is to examine whether humans can write better captions than models. The evaluation task is the appropriateness evaluation task, which consists of evaluating how appropriate the model-generated captions are versus ground-truth captions. Both tasks were performed by the same human subjects for the quality assessment of the data set.

\subsection{Figure Caption Generation Task}


\begin{table*}[!htb]
    \centering
    \begin{tabular}{lcrrrrrr}
        \hline 
        \textbf{Annotator} & \textbf{Inputs} &  \textbf{BLEU-4}& \textbf{METEOR}& \textbf{ROUGE-L} &  \textbf{SPICE} &\textbf{CIDEr}  \\
        \hline  
        1. Annotator 1 &Figure-only& 2.4 & 8.3 & 13.2 & 9.4 & 14.6 \\
        2. Annotator 2 &Figure-only&  \textbf{3.8} &  \textbf{10.1} &  \textbf{21.5} & 8.9 & \textbf{23.8} \\
        3. M4C-Captioner & Image and OCR features &  3.6 & 7.6 & 20.5 &  \textbf{11.5} & 18.7 \\
        \hline
        4. Annotator 1 &Figure-Mention& \textbf{7.7} & 13.4 & 19.1 & 15.9 & 11.3 \\
        5. Annotator 2 &Figure-Mention& 7.5 & \textbf{14.8} & 24.8 & 14.3 & 18.8 \\
        6. M4C-Captioner & Mention, Figure and OCR features &  5.5 & 11.6 & \textbf{28.1} & \textbf{16.1} &\textbf{47.7} \\
        \hline
    \end{tabular}
    \caption{Automatic evaluation scores on human-generated captions. The model has similar performances when the figure is the only available source. Using knowledge from vision and text modality, the model has a larger gain on CIDEr scores. } 
    \label{human_result_generation}
\end{table*}


\begin{table*}[!htb]
    \centering
    \begin{tabular}{lrrr}
        \hline
        \textbf{Model} & \textbf{Average Scores} & \textbf{Average Scores} & \textbf{Cohen-Kappa Scores} \\
         & \textbf{Evaluator 1} & \textbf{Evaluator 2} &  \\
        \hline
        1. M4C-Captioner (Mention, Figure and OCR features) & {1.8} &{2.13} & {0.27}  \\
        2. M4C-Captioner (Mention and OCR features) & \textbf{2.03} &\textbf{2.35} & {0.23}\\
        3. M4C-Captioner (Mention Only) & {1.8} &{2.22} & {0.31}\\
        4. M4C-Captioner (Figure and OCR features) &  {1.91} & {2.08} & {0.36} \\
        \hline
        5. Ground truth & {1.95} & {2.07} & {0.32} \\
        \hline
\end{tabular}
\caption{Average appropriateness score on model-generated and ground truth captions. Two evaluators gave low scores on both model-generated and ground truth captions, with the fair inter-annotator agreement.}
\label{appropriatness_score}
\end{table*}

The figure caption generation task is to generate captions under two conditions separately: 1. Figure-only: Human annotators write captions given only figures. This is to compare with captions generated by M4C-Captioner that only has access to figures and OCR features. 2. Figure-Mention: Human annotators write captions given both figures and their mention-paragraphs. We randomly selected 100 figures from the test set and to compare human-generated captions with captions generated by M4C-Captioner. 

The table~\ref{human_result_generation} shows automatic evaluation results for human caption generation tasks. Given only figures (rows \#1, 2), both annotators got low scores across all metrics, among those, annotator 2 led all metrics except SPICE. Since humans perform OCR naturally with their eyes we compare with M4C-captioner (Figure and OCR features). It has the best SPICE score, although it outperformed annotator 1 in 4 of 5 evaluation metrics, it achieved similar performance compared with annotator 2. This shows that without additional knowledge, humans aren't that better than machines.

However, given mention-paragraphs and figures (rows \#4, 5), compared to the figure-only condition, both annotators got improved scores in BLEU-4, METEOR, ROUGE-L, and SPICE but lower scores in CIDEr. Previous studies have shown that CIDEr is more reliable as an evaluation metric for caption generation, and the lowered CIDEr scores indicates that humans are likely to struggle with additional knowledge. On the other hand, having access to full features, M4C-captioner gained a significantly better CIDEr score compared to human annotators. The automatic evaluation results of the human generation tasks show the steep difficulty in writing figure captions close to ground truth. 

Even given mention-paragraphs, our annotator wrote captions with low scores across all standard image captioning evaluation metrics. We ascribe it as figure captions are highly subjective and require in-domain knowledge to write. Although our annotators are researchers, they cannot be professional in all knowledge existing in the computer science domain. Granted mention-paragraphs and OCR tokens as external knowledge sources, and with large-amount data training, the model can significantly outperform humans.  

\subsection{Appropriateness Evaluation}


This task evaluates the appropriateness of model-generated and ground-truth captions. We used the same set of 100 figures as in the figure caption generation task, and placed ground-truth captions and model-generated captions in random order. Then, human evaluators rank each caption to give appropriateness scores (1-4) to each caption. The evaluation scale: 1. Inappropriate: a caption does not match the figure, is not a sentence, is wrong, or is misleading. 2. Not sure: It is impossible to judge appropriateness solely from the figure. 3. Possible: A possible candidate that is incomplete but not wrong. 4. Appropriate: An informative caption that interprets the figure well. Since an appropriate figure caption should stand alone and readers should understand the messages the figure wants to represent without referring to the body text, we do not show mention-paragraphs to evaluators. 

Table~\ref{appropriatness_score} shows the results of the evaluations. Two evaluators gave low average scores to both model-generated captions and ground-truth captions. In addition, evaluators only reached fair agreements on scoring (0.23-0.36). Using the mention and OCR features (row \#2), gets the best human evaluation scores and this is in line with the corresponding score in Table~\ref{main_result} where it also achieves the best CIDEr performance, indicating that human evaluation is reliable despite the fair agreements. The evaluation results indicate that the model-generated and ground-truth captions are not always informative to both evaluators, which reveals the need to improve caption writing quality and model performance. We observed that captions tend to be written without following specific rules, and this may contribute to lack of agreement. With low inter-rater agreements, we found how informative a figure caption is highly subjective and depends on in-domain background knowledge evaluators have. 

\section{Related Work}

Unlike natural image captioning, figure captioning has been scarcely studied in history. SciCap~\cite{hsu-etal-2021-scicap-generating} is the most recent work on scientific figure captioning, they released a large-scale scientific figure captioning dataset that includes figures from academic papers in arXiv dataset. Before SciCap, FigCAP~\cite{chen2019figure}~\cite{Chen_2020_WACV} and FigureQA~\cite{kahou2017figureqa} are two figure captioning datasets, but their figures are synthesized. We decided to extend and study on SciCap dataset, since its figures are from real-world scientific papers. In this paper, we also have leveraged multimodal knowledge using pre-trained models. 

Multimodal machine learning is to model knowledge across various modalities. The closest multimodal task to figure captioning is image captioning, a popular architecture is encode-decoder, where the decoder learns to generate captions conditioned on visual features extracted from the encoder. Recent works on integrating texts in natural images for visual question answering and image captioning tasks are based on transformer architecture augmented with a pointer network~\cite{hu2020iterative, sidorov2019textcaps}. The transformer enriches representations by integrating knowledge from both text and visual modality. The pointer network dynamically selects words from the fixed dictionary or OCR tokens during generation.  

Using knowledge embedded in pre-trained models is a common practice in solving multimodal tasks. In this work, we used SciBert~\cite{beltagy-etal-2019-scibert}, a BERT model \cite{devlin-etal-2019-bert} that was pre-trained in scientific papers, to obtain informative representations for the texts extracted from computer science papers. Since terms that exist in the figures may be uncommon words, we also used FastText~\cite{Bojanowski2017EnrichingWV} to obtain word embeddings with subword information. For visual modality, we used Renst152~\cite{he2016deep} and Faster R-CNN~\cite{NIPS2015_14bfa6bb} used in extract features from images and bounding boxes.

\section{Conclusion}

In this paper, we study the challenges of the scientific figure captioning task. Extending from the previous study~\cite{hsu-etal-2021-scicap-generating}, we reframe this task as a knowledge-augmented image captioning task, that is, a model needs to use knowledge extracted across modalities to generate captions. To this end, we released a new version of the SciCap dataset: SciCap+ by augmenting figures with their mention-paragraphs and OCR tokens. We used M4C-Captioner model as the baseline model to utilize knowledge across three modalities: mention-paragraphs, figures, and OCR tokens. The automatic evaluation experiments further reveal that using knowledge significantly improves evaluation metric scores. Compared with human-generated captions, we found models can generate better captions than humans regarding the automatic evaluation metrics. However, human evaluations demonstrated that writing scientific figure captioning is challenging even for humans, and the model-generated figure captions, despite their reasonable automatic evaluation quality, are still far from achieving a level appropriate for humans. The release of the SciCap+ dataset is to promote the further development of scientific figure captioning. For future work, we are interested in how to use multimodal pretraining strategies in this task. 

\section{Acknowledgment}
These research results were partly obtained from the commissioned research (No. 225) by National Institute of Information and Communications Technology (NICT), Japan, and partly obtained from the first author's internship research under NICT.

\bibliography{anthology,custom_bib}

\appendix
\end{document}